\documentclass[11pt,a4paper]{article}
\usepackage[hyperref]{acl2020}
\usepackage{times}
\usepackage{latexsym}
\usepackage{graphicx}
\usepackage{float}
\graphicspath{ {images/} }

\usepackage{microtype}

\aclfinalcopy

\title{Machine-Generated Text Detection using Deep Learning}
\author{%
  \begin{tabular}{c}
    Raghav Gaggar\thanks{\ \ These authors contributed equally to this work.}\textsuperscript{*†} \hspace{1cm} Ashish Bhagchandani\footnotemark[1]\textsuperscript{*†} \hspace{1cm} Harsh Oza\footnotemark[1]\textsuperscript{*†} \\
    \normalfont\texttt{gaggar@usc.edu} \hspace{1cm} \normalfont\texttt{ashishna@usc.edu} \hspace{1cm} \normalfont\texttt{hoza@usc.edu}\\\\
    University of Southern California\textsuperscript{\dag}
  \end{tabular}
}

\begin{document}
\maketitle
\begin{abstract}
Our research focuses on the crucial challenge of discerning text produced by Large Language Models (LLMs) from human-generated text, which holds significance for various applications. With ongoing discussions about attaining a model with such functionality, we present supporting evidence regarding the feasibility of such models. We evaluated our models on multiple datasets, including Twitter Sentiment, Football Commentary, Project Gutenberg, PubMedQA, and SQuAD, confirming the efficacy of the enhanced detection approaches. These datasets were sampled with intricate constraints encompassing every possibility, laying the foundation for future research. We evaluate GPT-3.5-Turbo against various detectors such as SVM, RoBERTa-base, and RoBERTa-large. Based on the research findings, the results predominantly relied on the sequence length of the sentence.\\
\end{abstract}
\section{Introduction}

In recent years, the field of natural language processing (NLP) has witnessed revolutionary advancements, primarily fueled by the advent of large language models like OpenAI's ChatGPT \citep{brown2020language} \citep{radford2019language}. These models have demonstrated remarkable capabilities in generating human-like text, leading to widespread applications in various domains, including customer service, content creation, and education. However, the increasing sophistication of these models also poses significant challenges, particularly in distinguishing machine-generated text from human-generated content. The ability to make this distinction is crucial for maintaining the integrity and trustworthiness of digital communication.

This paper presents evidence which helps to support the challenge that human-generated sentences can be differentiated from sentences generated from GPT-3.5-Turbo. Leveraging state-of-the-art machine learning models such as RoBERTa-Base, RoBERTa-Large, and SVM, our approach aims to detect subtle differences in language patterns, stylistic features, and semantic nuances. Previous work in this area has primarily focused on identifying generic characteristics of AI-generated text, often overlooking the specific attributes of advanced models like ChatGPT \citep{chakraborty2023possibilities}. Our work builds upon these foundations but introduces a targeted analysis aligned with the unique linguistic features of ChatGPT-generated content.

To develop and evaluate our model, we compiled a diverse dataset comprising thousands of sentences from human sources in domains including Sports, Medical, Twitter reviews, Text comprehension, and Literature spanning various genres and styles. Further, using these sentences to generate similar sentences from the GPT-3.5-Turbo API of ChatGPT. This dataset was used in the pre-processing stage of natural language embeddings and feature extraction before using it for any classification model.

The significance of this research lies not only in its immediate application for content verification but also in its broader implications for the field of digital forensics and the ethics of AI in communication. As language models continue to evolve, distinguishing between human and AI-generated content will become increasingly challenging, necessitating ongoing research and development in this area \citep{crothers2023machine}.

In this paper, we describe our methodology, the architecture of our detection model, and the results of our experiments. We also discuss the broader implications of our findings in context with sentence features and patterns.

\section{Related Work}

Numerous investigations have been undertaken to discern the features differentiating human-generated text from machine-generated text. Let's delve into each of these approaches.

Recent LLMs like ChatGPT can generate text to compose essays, describe art in detail, create AI art prompts, have philosophical conversations, and even code for you. To detect such intricate patterns, a new methodology was proposed in DetectGPT, where they discovered how LLMs operate where the models tend to create text that falls into specific patterns, particularly in regions where the model's calculations show negative curvature. DetectGPT utilizes log probabilities from the LLM and random perturbations that it generates to determine whether the text is machine-generated \citep{mitchell2023detectgpt}. It is particularly good at identifying fake news articles made by models like GPT-NeoX and outperforms most zero-shot methods. They test LLM performance with six diverse datasets: XSum for fake news detection, SQuAD for academic essays, Reddit WritingPrompts for creative writing, WMT16 in English and German, and PubMedQA for long-form expert answers. The future work that authors in DetectGPT wanted to explore was to see how watermarking algorithms worked with detection algorithms like DetectGPT.

\citep{guo2023close} proposed an HC3 (Human ChatGPT Comparison Corpus) dataset, which consists of nearly 40K questions and their corresponding human/ChatGPT answers, and developed a ChatGPT detection model to differentiate human and Chatgpt generated text. The detector is primarily created by fine-tuning RoBERTa model on the dataset, and the authors propose two methods for training it. The first method uses only the pure answered text, while the second method utilizes the question-answer text pairs for joint model training.

In 2019, \citep{solaiman2019release} built a detector for GPT2-generated output by fine-tuning RoBERTa using outputs from the largest GPT2 Model, which was with 1.5 billion parameters and was capable of detecting if the text was a machine-generated text or not. They conducted this research by using RoBERTa-base with 125 million parameters and RoBERTa-large with 356 million parameters as the foundation for their sequence classifier. RoBERTa, distinct from GPT-2 in terms of architecture and tokenizer, is a masked and non-generative language model. They achieved an accuracy of approximately 95\% for this research.

\citep{kirchenbauer2023watermark} proposed a watermarking framework for language models. Watermarking basically means embedding signals in the machine-generated text so it is not detected by the human eye, but detected algorithmically and helps to decrease potential harm from these LLM models. Kirchenbauer tested this technique on an Open Pretrained Transformer (OPT) model with multi-billion parameters. The authors used an approach of categorizing watermark tokens into green and red lists for distinct patterns. This approach helps in use cases like plagiarism check and copyright protection. To mimic diverse language modeling situations in their dataset, they extracted random text portions from the news-like section of the C4 dataset \citep{raffel2020exploring}. Each extracted string has a fixed number of tokens removed from the end, forming a baseline completion, while the rest of the tokens serve as the prompt.

On the other hand, on social media applications like Twitter, which are vulnerable to generating misinformation and fake news, \citep{kumarage2023stylometric} authors proposed a novel stylometric detection algorithm to detect AI-generated tweets that are human-like. Authors use BERT and ensemble approach, incorporating stylometric features as their baseline study. they test their models on two datasets: one we made to mimic human-to-AI author changes on a Twitter user's timeline and the publicly available TweepFake dataset. They use stylometric features to analyze text for stylistic signals, categorizing them into phraseology, punctuation, and linguistic diversity. Using this with a model like RoBERTa which classifies on top of the extracted features, they found that this method performs the best with XGBoost compared to logistic regression and random forest classifiers.

There are also tools that are developed, like \citep{gptzero} which is a tool used for analyzing text, employing two primary metrics, namely perplexity and burstiness, to differentiate between machine-generated and human text. It offers a publicly accessible API that generates a confidence score indicating the likelihood of a given text being machine-generated or not.

We plan to investigate the potential impact of different data sources and types of machine-generated text on the AUCROC of our model. It may help us understand the limitations of our approach and identify areas for future research. By building on the existing work in this field and extending it in these ways, we hope to contribute to developing more accurate and effective methods for distinguishing between human-generated and machine-generated text (ChatGPT). The model can provide a valuable addition to the literature on identifying fake and synthetic data, as it addresses a specific type of data generation method and can help improve the accuracy of existing methods. To do this, we build a custom dataset by combining multiple types of text like PubMed data, SQuAD, Twitter feed data, Football commentary, and novels.\\

\section{Proposed Approach}
This section presents a nuanced and multi-faceted approach for discerning text's origin, specifically focusing on the distinction between human-generated and ChatGPT-generated sentences. This requires training and fine-tuning state-of-the-art NLP models on a diverse dataset of human-generated text and ChatGPT-generated text. The ultimate goal is to enable effective moderation and filtering of text content across different platforms and applications, thus ensuring the safety and integrity of online communication.\\
The models should be able to accurately identify text generated by ChatGPT across multiple domains, including Sports, Medical, Twitter reviews, Text comprehension, and Literature. The research aims to provide insights into the decision-making process of the models and the characteristics of ChatGPT-generated text that distinguish it from human-generated text.

\subsection{Dataset Description}
The dataset used in this study is a combination of five different sources of human-generated text, including Twitter Sentiment, Football Commentary, Project Gutenberg, PubMedQA, and SQuAD datasets. The purpose of using these datasets is to improve the model's ability to understand and analyze both short-form and long-form text and complex medical jargon. 

We split the Football Commentary dataset into 100-word and 150-word sentences, truncated SQuAD questions to a maximum of 200 words, and extracted 200-word sentences from the Project Gutenberg dataset to form custom literature sentences.

\begin{figure*}[t]
    \centering
    \includegraphics[width=\textwidth]{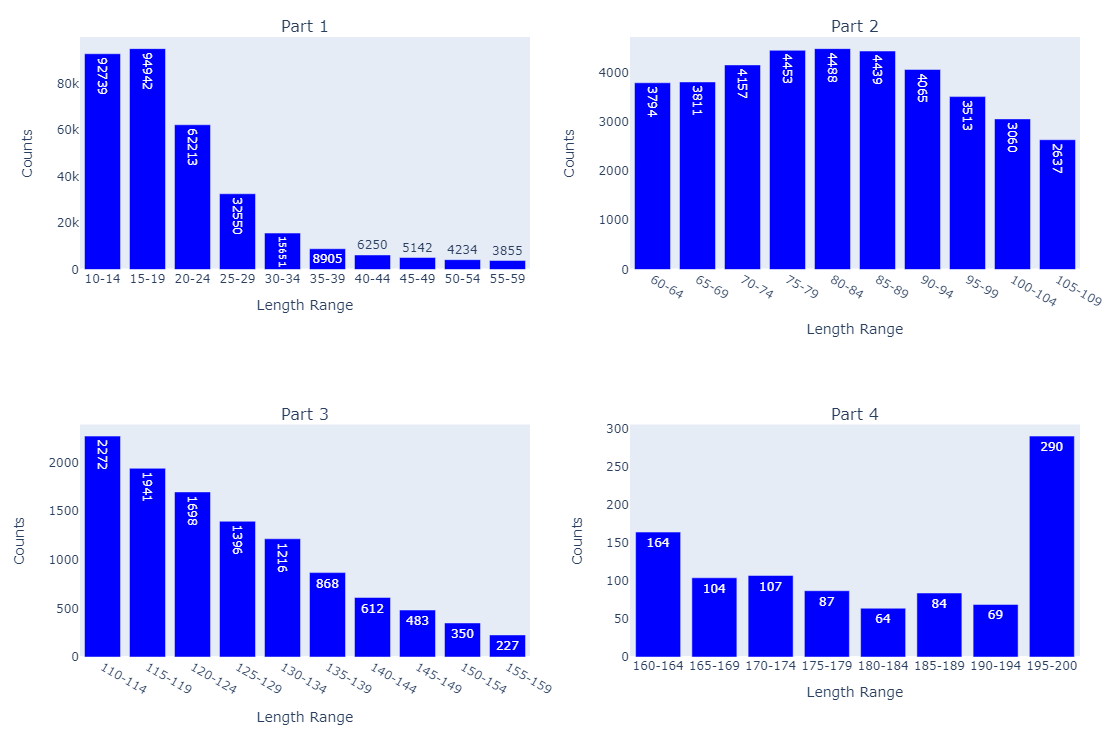}
    \caption{Dataset distribution across different range of lengths}
    \label{fig:newplot}
\end{figure*}

The merged human-generated text resulted in 2,534,498 sentences, which were grouped into categories of 10-200 words with 5-word increments. Normalization was performed to ensure 16\% of sentences in each category, resulting in a final dataset of 400,015 sentences as shown in figure \ref{fig:newplot}.

To generate machine text, we rephrased the selected sentences using the GPT-3.5-turbo chatgpt API from OpenAI, with a prompt to ensure that the length of the rephrased sentence is the same as the original sentence as shown in Table \ref{tab:rephrase}. The total number of rows in our dataset is 800,030, consisting of both human and machine generated text, with sentences ranging from 10 to 200 words in length. We truncated sentences longer than 200 words and removed those with fewer than 10 words.

\begin{table}[hbt!]
\centering
\begin{tabular}{lp{3.5cm}}
\hline
\textbf{Prompt} & \textbf{ChatGPT Response}\\
\hline
Please rephrase this  & Rephrased sentence \\
sentence make sure the & from Chatgpt\\
words length is equal to \\
the given sentence.\\
  \\
\hline
\end{tabular}
\caption{Example Generating Rephrase Data.}\label{tab:rephrase}
\end{table}

\subsection{Model Description}
Our methodology encompasses three distinct cases, each tailored to leverage the strengths of a specific model. In the first case, we employ Support Vector Machines (SVM) to conduct a meticulous analysis, followed by using RoBERTa-base in the second case and RoBERTa-large in the third case.

The models should be able to accurately identify text generated by ChatGPT across multiple domains, including Sports, Medical, Twitter reviews, Text comprehension, and Literature. The research aims to provide insights into the decision-making process of the models and the characteristics of ChatGPT-generated text that distinguish it from human-generated text.

\subsubsection{SVM}
In the case of the SVM classifier, a powerful tool in machine learning, we employed a radial basis function (RBF) kernel to discern patterns within the feature space. The RBF kernel is known for capturing complex relationships in non-linear data, making it particularly well-suited for distinguishing between human-generated and ChatGPT-generated sentences. We used the TF-IDF (Term Frequency-Inverse Document Frequency) vectorization technique to represent the textual data. TF-IDF accounts for the frequency of terms within each document and considers the importance of terms in the broader corpus. This combination of the RBF kernel and TF-IDF vectorization ensures that our SVM model is equipped to handle the intricacies of natural language, capturing both semantic nuances and contextual variations for accurate and robust classification. 

The choice of the SVM classifier with an RBF kernel and TF-IDF stems from its status as a well-established and widely-used baseline approach in the domain of text classification. This approach has been successfully employed in previous text classification studies, as demonstrated by its use in studies such as \citep{das2018improved}. 

\begin{figure*}[t]
    \centering
    \includegraphics[width=\textwidth]{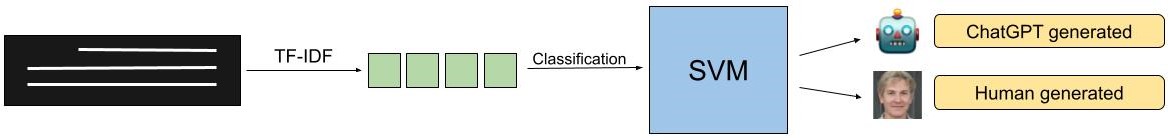}
    \caption{SVM Pipeline.}
    \label{fig:svm}
\end{figure*}

\begin{figure*}[t]
    \centering
    \includegraphics[width=\textwidth]{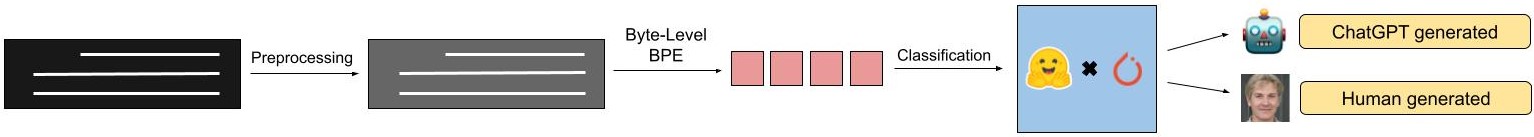}
    \caption{RoBERTa Pipeline.}
    \label{fig:roberta}
\end{figure*}

It serves as a benchmark against which the performance of more complex models can be evaluated. In contrast to the RoBERTa models, the SVM-RBF-TF-IDF approach is lightweight, requiring less time and computational resources for both training and inference. This characteristic renders it more amenable to scenarios with resource constraints. Additionally, the SVM model proves to be robust, showcasing competitive performance compared to other machine learning models of similar size and complexity in the realm of text classification. 

However, it is essential to note that while the SVM model offers efficiency and robustness, it does not attain state-of-the-art performance levels achieved by the more advanced RoBERTa models, as detailed in the subsequent sections of our analysis.

\subsubsection{RoBERTa-base}
In our second case, we employed the RoBERTa-base model, a transformer-based architecture renowned for its exceptional natural language processing capabilities. To tailor this pre-trained language model to our specific binary classification task—distinguishing between human-generated and ChatGPT-generated sentences we augmented RoBERTa-base with a fully connected (FC) layer followed by a sigmoid activation function at the end. This modification allowed us to transform the contextualized representations learned by RoBERTa into a binary classification decision. The RoBERTa model, with its deep bidirectional architecture and extensive pre-training on vast corpora, captures intricate linguistic patterns and semantic relationships, enhancing our ability to discern nuanced differences in text origins. Adding the FC+sigmoid layer enables the model to map these representations to a binary decision space, facilitating accurate classification. This hybrid approach leverages the strengths of transformer-based language models while tailoring them to the specific demands of our binary classification task.

The RoBERTa-base model, augmented with an (FC+sigmoid) layer, is chosen for its excellence in NLP tasks, supported by its deep bidirectional architecture. It showcases versatility and is successfully utilized in text detection tasks by studies like \citep{mitchell2023detectgpt} and [DistilBERT]. With 125 million parameters, RoBERTa-base balances complexity and efficiency compared to RoBERTa-large (355 million parameters). This choice facilitates an exploration of model size-performance trade-offs, offering valuable insights into transformer-based models for binary classification tasks.

The RoBERTa-base model, enhanced with an (FC+sigmoid) layer, proves effective in our binary classification, nearing state-of-the-art performance observed in similar studies. Its adeptness in capturing intricate linguistic nuances is countered by its larger size (125 million parameters), demanding more training and inference resources. While excelling in accuracy, its computational complexity introduces challenges in efficiency when compared to the lightweight SVM model. This highlights the nuanced trade-offs between model sophistication, performance, and resource requirements within the context of our classification task.gths of transformer-based language models while tailoring them to the specific demands of our binary classification task.

\subsubsection{RoBERTa-large}
In our third case, we employed the formidable RoBERTa-large model, a transformer-based architecture renowned for its extensive depth and superior natural language understanding capabilities. Augmented with an additional fully connected (FC) layer and a sigmoid activation function at the end, RoBERTa-large was tailored for our binary classification task, distinguishing between human-generated and ChatGPT-generated sentences. With a substantial parameter count of 355 million, RoBERTa-large surpasses its base counterpart in both depth and complexity. 

RoBERTa-large, enhanced with an appended (FC+sigmoid) layer, represents a powerful extension of its base counterpart and proves instrumental in our binary classification task. Acknowledged as a highly proficient transformer for NLP tasks, this model, with its extensive depth and attention mechanisms, has been a preferred choice in recent studies addressing machine-generated text detection tasks, as demonstrated by its implementation in studies such as \citep{mitchell2023detectgpt} and [DistilBERT]. The RoBERTa-large model, with its 355 million parameters, boasts an intricate understanding of contextual relationships within textual data, enhancing its ability to discern subtle distinctions between human-generated and ChatGPT-generated sentences. Notably, it has exhibited outstanding performance in similar studies, achieving state-of-the-art results on our dataset. 

However, the heightened complexity and larger size of RoBERTa-large introduce computational challenges, requiring substantial training and inference resources. This trade-off between enhanced capabilities and increased computational demands prompts a nuanced consideration of its suitability for our specific classification task, especially in comparison to the more lightweight SVM and RoBERTa-base models. \\

\section{Experimental Setup}
This section outlines the experimental configuration for both the dataset and the model.

\subsection{Dataset Experiments}
For our experimental approach, we organized our datasets into distinct training, testing, and validation sets, each comprising sentences ranging in length from 10 to 200 words. To ensure a granular examination of model performance across different sentence lengths, we further divided the data within each set into specific ranges, such as 10-14 words, 15-19 words, and so on. 

This stratification allowed us to scrutinize the models' proficiency in handling varying sentence lengths, offering insights into their adaptability across a spectrum of linguistic contexts. By systematically categorizing the data based on sentence length ranges, our experimental design aimed to comprehensively evaluate model robustness and effectiveness in capturing nuances across diverse sentence lengths.

\subsection{Model Experiments}
Furthermore, to comprehensively assess the discriminatory power of each model across different sentence length ranges, we recorded the Area Under the Receiver Operating Characteristic curve (AUC-ROC) for each specific range and the cumulative performance across all ranges. 
This recording strategy allowed us to capture the variations in model performance at distinct sentence lengths and derive an aggregate measure of effectiveness. 

The AUC-ROC analysis is a robust metric, offering a holistic understanding of each model's ability to discriminate between human-generated and ChatGPT-generated sentences within the specified length categories. Combining results across all ranges, our evaluation framework thoroughly examines the models' overall discriminatory performance, contributing valuable insights to sentence-length-specific classification.\\

\section{Results and Discussion}

\begin{figure*}[hbtp]
    \centering
    \includegraphics[width=0.8\textwidth]{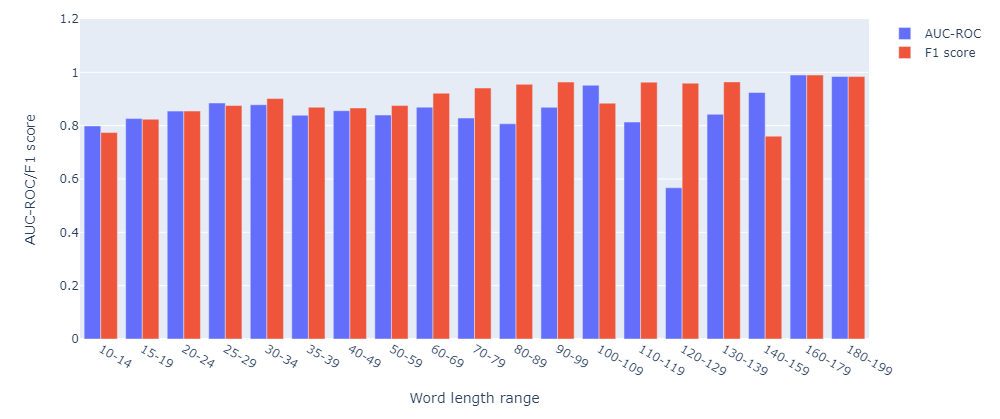}
    \caption{Performance of SVM over texts of different lengths.}
    \label{fig:svm_res}
\end{figure*}

\begin{figure*}[hbtp]
    \centering
    \includegraphics[width=0.8\textwidth]{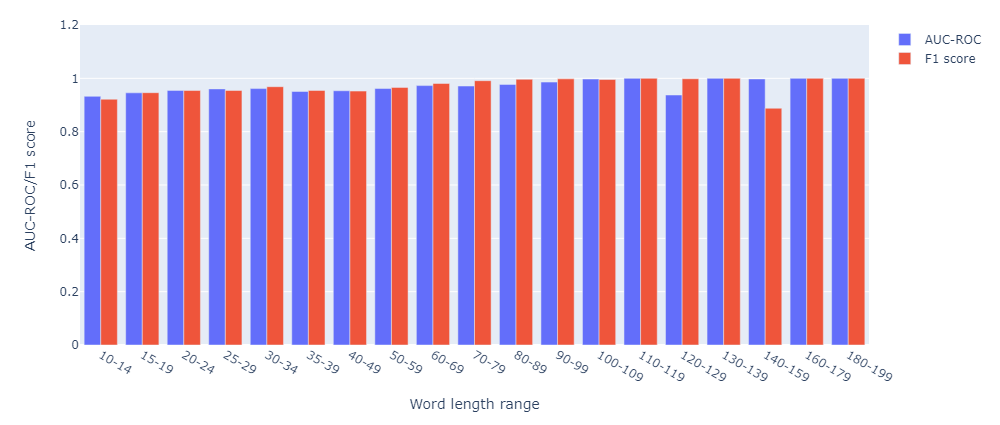}
    \caption{Performance of RoBERTa-base over texts of different lengths.}
    \label{fig:roberta_base_res}
\end{figure*}

\begin{figure*}[hbtp]
    \centering
    \includegraphics[width=0.8\textwidth]{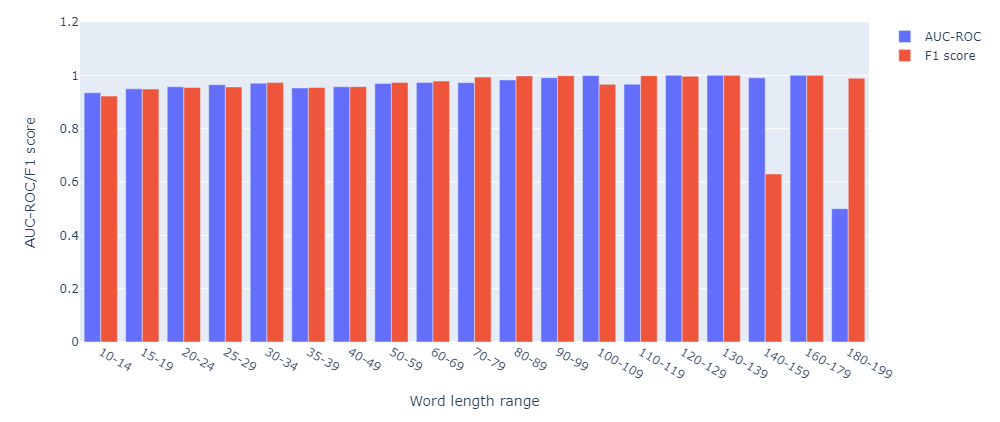}
    \caption{Performance of RoBERTa-large over texts of different lengths.}
    \label{fig:roberta_large_res}
\end{figure*}

In our first experimental results in table \ref{tab:models}, we present the AUC-ROC scores for different models across various datasets, including training (Train), testing (Test), and validation (Val) sets. The SVM model demonstrates strong discriminatory performance with AUC-ROC scores on the Train, Test, and Val sets. Moving to transformer-based models, RoBERTa-base exhibits elevated performance compared to SVM, while RoBERTa-large further enhances this performance across the Train, Test, and Val sets. These scores provide a quantitative overview of each model's effectiveness in distinguishing between human-generated and ChatGPT-generated sentences across different datasets, highlighting the robust discriminatory capabilities of the RoBERTa models.

\begin{table}
\centering
\begin{tabular}{lcccc}
\hline
\textbf{Model} & \textbf{Train} & \textbf{Test} & \textbf{Val}\\
\hline
SVM & 91.83\% & 84.81\% & 84.73\% \\
RoBERTa-base & 96.85\% & 95.24\% & 95.53\% \\
RoBERTa-large & 97.11\% & 95.53\% & 95.76\% \\
\hline
\end{tabular}
\caption{AUC-ROC comparison for models.}\label{tab:models}
\end{table}

\begin{table}
{\fontsize{10}{14}\selectfont
\centering
\begin{tabular}{lcccc}
\hline
\textbf{Range} & \textbf{SVM} & \textbf{RoBERTa-} & \textbf{RoBERTa-} \\
                        &               & \textbf{base}       & \textbf{large}      \\
\hline
10-14 & 0.799119 & 0.932474 & 0.934662\\
15-19 & 0.826536 & 0.945634 & 0.949327\\
20-24 & 0.854538 & 0.954775 & 0.957018\\
25-29 & 0.884957 & 0.960358 & 0.964959\\
30-34 & 0.87853 & 0.96152 & 0.970518\\
35-39 & 0.839002 & 0.95069 & 0.952666\\
40-49 & 0.85617 & 0.953383 & 0.956988\\
50-59 & 0.839941 & 0.96173 & 0.969639\\
60-69 & 0.869463 & 0.972758 & 0.973368\\
70-79 & 0.82854 & 0.971184 & 0.972567\\
80-89 & 0.807112 & 0.977258 & 0.982704\\
90-99 & 0.868776 & 0.985965 & 0.991071\\
100-109 & 0.951641 & 0.997299 & 0.999021\\
109-119 & 0.813498 & 1 & 0.966667\\
120-129 & 0.567105 & 0.9375 & 1\\
130-139 & 0.842593 & 1 & 1\\
140-159 & 0.924376 & 0.997717 & 0.990476\\
160-179 & 0.99 & 1 & 1\\
180-199 & 0.984375 & 1 & 0.5\\
\hline
\end{tabular}
}
\caption{AUCROC scores achieved w.r.t ranges.}\label{tab:ranges}
\end{table}

For our second experiment, we recorded AUC-ROC scores for each range length of sentences across three different models: SVM, RoBERTa-base, and RoBERTa-large. The table \ref{tab:ranges} displays the AUC-ROC scores for each model within specific sentence length ranges.

As sentence lengths increase, RoBERTa models consistently outperform SVM, showcasing robustness in capturing nuanced patterns. In table \ref{tab:rangesf1}  RoBERTa-large maintains competitive F1 scores even in longer sentence ranges, indicating its proficiency across diverse linguistic contexts. These results offer valuable insights into the models' efficacy at different scales of textual complexity, emphasizing the performance variations observed across various sentence length intervals.

These scores provide a detailed breakdown of each model's performance within distinct sentence length intervals, offering valuable insights into their discriminatory abilities across varying linguistic contexts.\\

\begin{table}
{\fontsize{10}{14}\selectfont
\centering
\begin{tabular}{lccccc}
\hline
\textbf{Range} & \textbf{SVM} & \textbf{RoB-base} & \textbf{RoB-large} \\
\hline
10-14 & 0.774232 & 0.921676 & 0.922112\\
15-19 & 0.824198 & 0.94617 & 0.948374\\
20-24 & 0.854733 & 0.9543 & 0.954323\\
25-29 & 0.875496 & 0.954439 & 0.956495\\
30-34 & 0.901613 & 0.968602 & 0.972937\\
35-39 & 0.869093 & 0.95452 & 0.954179\\
40-49 & 0.866435 & 0.952179 & 0.957542\\
50-59 & 0.875376 & 0.966087 & 0.973392\\
60-69 & 0.921739 & 0.980781 & 0.978584\\
70-79 & 0.941176 & 0.99111 & 0.993544\\
80-89 & 0.954955 & 0.996504 & 0.997619\\
90-99 & 0.963504 & 0.998038 & 0.998717\\
100-109 & 0.884438 & 0.995527 & 0.9665\\
109-119 & 0.963151 & 1 & 0.998805\\
120-129 & 0.959459 & 0.998361 & 0.996774\\
130-139 & 0.9642 & 1 & 1\\
140-159 & 0.760638 & 0.887805 & 0.630081\\
160-179 & 0.989899 & 1 & 1\\
180-199 & 0.984127 & 1 & 0.989247\\
\hline
\end{tabular}
}
\caption{F1 scores achieved w.r.t ranges.}\label{tab:rangesf1}
\end{table}

\section{Conclusion and Future Work}
In our research, we effectively create a well-constructed labeled dataset encompassing texts originated by both humans and ChatGPT (GPT-3.5 turbo) from five distinct sources. The sentences in the dataset exhibit a varied length range, spanning from 10 to 200 words. Additionally, we have designed and trained multiple classifiers to differentiate between texts generated by humans and ChatGPT.

In the future, when creating datasets, we can aim to add a wider range of text from more sources. This approach will make the data more versatile, suitable for different fields, and capable of handling texts of different lengths. To improve the results of classification, progress can be achieved by developing and training more advanced models. Considering the increasing development of Language Model Models (LLMs), our work can be expanded to incorporate all noteworthy LLMs, such as Llama, Orca, Falcon, and Palm, for both dataset creation and detection purposes.
For further model improvement, we can design a system to identify phrases within document sentences, aiming to distinguish between machine-generated and human-generated text. Finally, exploring zero-shot and one-shot learning systems like \cite{mitchell2023detectgpt} can eventually aid in saving resources and time needed for training complex classifiers.

\nocite{*}
\bibliography{acl2020}
\bibliographystyle{acl_natbib}

\end{document}